\documentclass[runningheads]{llncs}

\usepackage{eccv}

\usepackage{eccvabbrv}

\usepackage{graphicx}
\usepackage{booktabs}
\usepackage[table]{xcolor}
\usepackage{multirow}
\usepackage[accsupp]{axessibility}  

\usepackage{hyperref}

\usepackage{orcidlink}
\newcommand\samethanks[1][\value{footnote}]{\footnotemark[#1]}
\newcommand{\sys}{DINOde}

\begin{document}

\title{DINOde: Continuous Vision-Text Alignment for \\Open-Vocabulary Semantic Segmentation} 

\titlerunning{DINOde: Continuous Vision-Text Alignment for OVSS}

\author{Sung-Hoon Yoon\inst{1}\orcidlink{0000-0001-5851-2031}\thanks{Equal contribution.} \and
Hoyong Kwon\inst{2}\orcidlink{0000-0002-5888-1002}\samethanks \and
Changgyoon Oh\inst{2}\orcidlink{0000-0003-4603-0486}\samethanks \and \\
Kuk-Jin Yoon\inst{2}\orcidlink{0000-0002-1634-2756}}

\authorrunning{Yoon et al.}

\institute{
Multimodal Intelligence and Perception Lab., DGIST, Republic of Korea\\
\email{shyoon@dgist.ac.kr}\\ 
\and
Visual Intelligence Lab., KAIST, Republic of Korea \\
\email{\{kwonhoyong3, changgyoon, kjyoon\}@kaist.ac.kr}
}

\maketitle
\begin{abstract}
Open-vocabulary semantic segmentation (OVSS) leverages textual semantics to segment objects beyond predefined categories. While the self-supervised model DINOv3 provides strong structured visual representations, its lack of native textual alignment hinders direct application to OVSS. To bridge this gap, we propose \sys, an ODE-based framework that continuously aligns CLIP text embeddings to the DINO visual manifold. Our approach employs two complementary components: (i) \textit{Semantic Text Flow} (STF), which evolves text embeddings toward the DINO manifold through a continuous ODE trajectory, and (ii) \textit{Global Context Flow} (GCF), which progressively refines the holistic image representation carried by DINO’s \texttt{CLS} token. To preserve the hyperspherical geometry of the feature space during this evolution, we further introduce \textit{Velocity Tangent Projection}, which constrains the learned velocity field to the tangent space through projection. By modeling alignment as a continuous trajectory, \sys~avoids the manifold entanglement inherent in discrete MLP projections and yields more robust cross-modal alignment. Extensive experiments demonstrate that \sys~consistently outperforms existing methods and achieves state-of-the-art performance across multiple OVSS benchmarks.
The code is available at \url{https://github.com/yoon307/DINOde}.
\end{abstract}

\section{Introduction}
\label{sec:intro}

Semantic segmentation~\cite{ocrnet, segformer, unet, maskformer, deeplabv3p, cffm, cffmpp, kim2026bootstrapping} has become a cornerstone task in computer vision, ranging from autonomous systems to medical image analysis. 
However, real-world deployment remains challenging due to the inherent open nature of the environment, where unseen categories frequently emerge. Consequently, recent research has pivoted toward open-vocabulary semantic segmentation (OVSS), which aims to segment unseen classes.

The emergence of large-scale vision-language pretraining such as the Contrastive Language-Image Pre-training (CLIP~\cite{CLIP}) model has significantly advanced this direction by providing a robust alignment between visual features and textual semantics.
Early works such as LSeg~\cite{LSeg2022} and TCL~\cite{tcl} employ static contrastive losses to establish direct pixel-text correspondences, while others like MaskCLIP~\cite{maskclip23} extract dense pseudo-masks directly from frozen encoders for training-free inference. More recently, frameworks have introduced hierarchical region-level grouping as seen in GroupViT~\cite{groupvit22}, or integrated specialized mask-adapted segmentation modules like OVSeg~\cite{ovseg23} and FC-CLIP~\cite{yu2023convolutions} to enhance spatial precision.
However, CLIP’s visual encoder is primarily optimized for global image-text alignment, often yielding coarse and spatially entangled representations that are suboptimal for dense prediction tasks. 

\begin{figure}[t]
  \centering
   \includegraphics[width=0.9\linewidth]{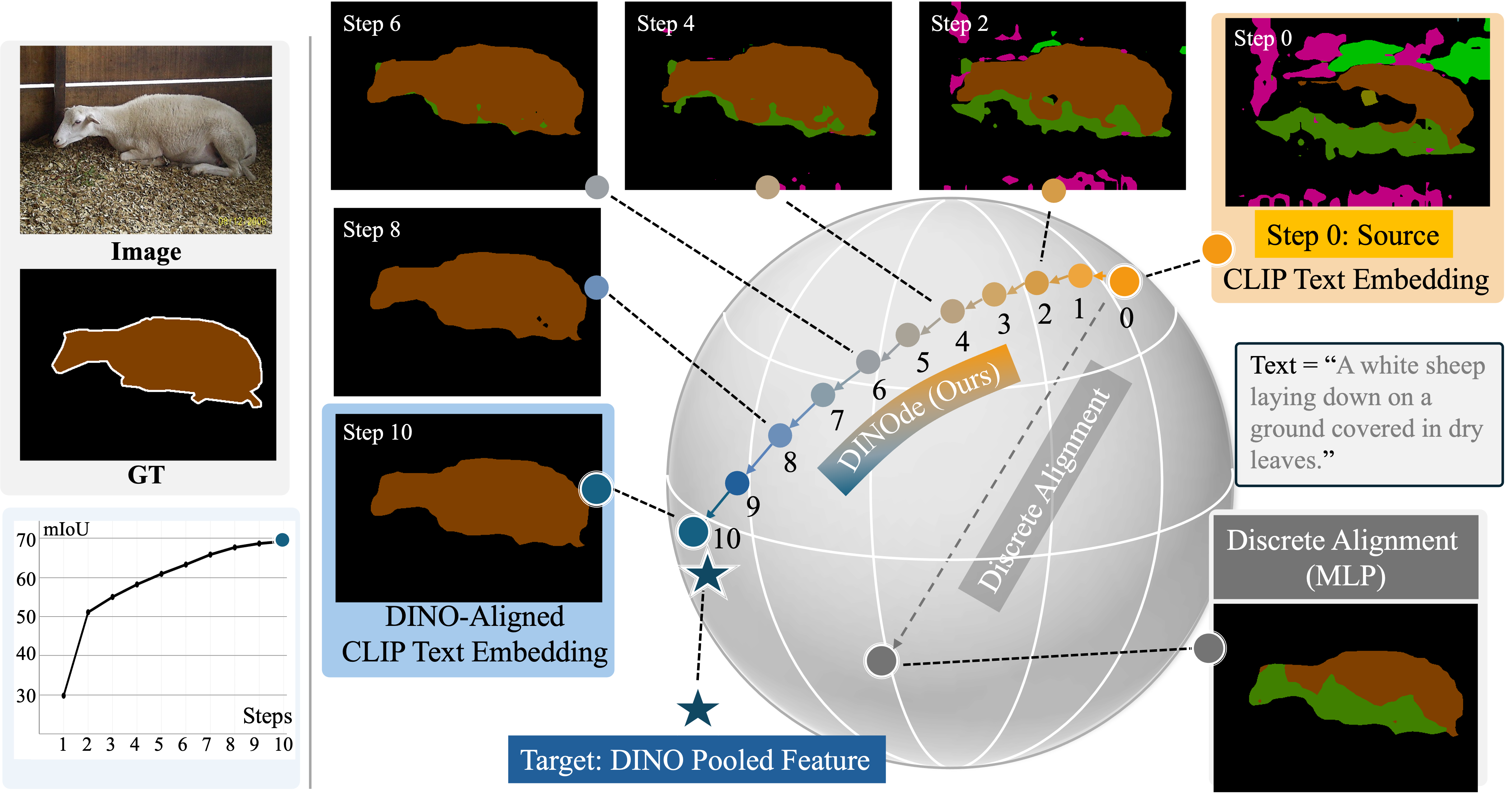}
   \caption{\textbf{Overview of the DINOde framework for OVSS}. We propose a continuous alignment strategy that bridges the gap between text embeddings and DINO visual features. The illustration depicts how a text embedding is progressively transformed into a DINO-aligned text embedding through an ODE-based trajectory on the unit sphere, resulting in step-by-step refinement of the semantic segmentation.}
   \label{fig:introduction}
\end{figure}

Meanwhile, self-supervised Vision Transformers (ViT) trained purely on visual data, such as DINO~\cite{dinov1} and DINOv2~\cite{dinov2}, exhibit superior localization capabilities and have thus gained increasing attention in dense prediction tasks. Nevertheless, as these self-supervised models operate exclusively within the visual modality, their representation spaces remain unaligned with textual semantics, rendering them incapable of direct open-vocabulary reasoning.

In line with this, the recent work \textit{dino.txt}~\cite{dinotxt} aligns a frozen DINOv2 backbone with a newly trained text encoder for open-vocabulary segmentation via locked image-text alignment. However, it requires training a text encoder from scratch on large-scale paired data and demands extensive computational resources for vision-text alignment. Another concurrent work, Talk2DINO~\cite{talk2dino}, leverages DINOv2’s self-attention heads to selectively align the most relevant visual regions with textual concepts. While this approach aligns DINOv2 spatial tokens with CLIP text embeddings without fine-tuning the backbones, it simply bridges the two feature spaces using a non-linear mapping function.

Recently, DINOv3~\cite{dinov3} has enabled the preservation of high-quality spatially detailed features through its Gram anchoring regularizer. Notably, simply aligning DINOv3 with the CLIP text embedding space using an MLP provides a surprisingly strong baseline with competitive performance. While aligning two frozen modalities is efficient and straightforward to optimize, MLP-based schemes that attempt to approximate this complex cross-modal transformation through a single-step instantaneous mapping often fail to preserve the topological relationships (e.g., semantic proximity between `cat' and `dog') between the representation manifolds.
We experimentally observe that MLP-based alignment leads to suboptimal performance, primarily because it adopts a manifold-agnostic Euclidean shortcut that disregards the intrinsic curvature of the data. This structural limitation induces semantic entanglement, wherein the neighborhood relationships from the original space are distorted within target space.

To address these limitations, we introduce \sys, a continuous ordinary differential equation (ODE)-based vision-text alignment framework. Instead of enforcing a rigid one-step projection, our model learns a smooth trajectory that gradually transforms the textual manifold into DINOv3’s representations, as shown in Fig.~\ref{fig:introduction}. 
Furthermore, by incorporating tangent projection on the hypersphere, we ensure the learned velocity field strictly adheres to geometric constraints, enabling a smooth and manifold-preserving flow. Here, we align both text embeddings and DINOv3's \texttt{CLS} token through separate ODE trajectories, maximizing global-local semantic coherence.
As a result, \sys~demonstrates state-of-the-art OVSS performance with only image-text pairs.

The contributions of our method can be summarized as follows:
\begin{itemize}
    \item We introduce \textbf{\sys}, a novel ODE-based alignment framework that continuously transforms  CLIP text embeddings toward DINOv3 features, ensuring a smooth and geometry-preserving vision-text mapping.
    \item By progressively aligning both text embeddings and DINOv3 \texttt{CLS} tokens through an ODE-based framework, the proposed DINOde effectively integrates both local and global information.
    \item We introduce Velocity Tangent Projections to enable geometric-constrained and manifold-preserving ODE learning.
    \item Extensive experiments across multiple benchmarks and unseen categories demonstrate the effectiveness and generalizability of our approach for open-vocabulary semantic segmentation.
\end{itemize}

\section{Related Work}
\label{sec:related}

\subsection{VLM-Based Open-Vocabulary Segmentation}
\label{subsec:vlm_ovss}

Foundational works in Open-Vocabulary Semantic Segmentation (OVSS) predominantly rely on VLMs, such as CLIP~\cite{CLIP}, for both visual and textual feature extraction. These approaches generally fall into two categories: decoder-free alignments and conditional decoder frameworks.

\subsubsection{Conditional Decoder Frameworks}
\label{subsubsec:vlm_decoder}
To address the severe localization deficiency of VLM-only features, fully-supervised conditional frameworks~\cite{LSeg2022, ghiasi2022scaling, xu2022simple, ovseg23, xdecoder23, xu2023side, yu2023convolutions, xu2023open, zhang2023simple} integrate the VLM with heavy segmentation decoders trained on base-class pixel annotations. For instance, LSeg~\cite{LSeg2022} appends a dense prediction transformer to the visual encoder to establish direct per-pixel correspondence with textual embeddings, while OVSeg~\cite{ovseg23} fine-tunes a complex Mask2Former-based decoder to generate class-agnostic mask proposals that are subsequently classified by a mask-adapted CLIP~\cite{CLIP}. This highlights an ongoing trade-off between fine-grained localization quality and annotation cost, motivating complementary research into more lightweight paradigms.

\subsubsection{Decoder-Free VLM Alignment}\label{subsubsec:vlm_align}
To avoid heavy, task-specific decoders, researchers often extract dense predictions directly from the VLMs. The training-free scheme~\cite{maskclip23, shin2022reco, bousselham2024grounding, wysoczanska2024clip, clearclip, sun2024clip, luo2024emergent, kang2024defense, proxyclip, sun2025cliper} leverages off-the-shelf VLMs without any additional training phase or parameter updates. For instance, MaskCLIP~\cite{maskclip23} alters the self-attention topology of the frozen CLIP~\cite{CLIP} visual encoder to extract dense pixel-level features directly, while FreeDA~\cite{freeda} enhances training-free matching by generating diffusion-augmented visual prototypes to enrich semantic representations. 
%
Alternatively, similar to weakly-supervised semantic segmentation studies that avoid dense pixel annotations~\cite{unlockingwsss, digwsss, ctiwsss, pcsswsss, evwsss, acrwsss, mctformerwsss, climswsss}, the weakly-supervised scheme in OVSS~\cite{groupvit22, luo2023segclip, tcl, ovseg23, mukhoti2023open, yi2023simple, cai2023mixreorg} relies solely on image-text pairs to align features without pixel-level annotations.
Notably, GroupViT \cite{groupvit22} learns hierarchical patch groupings by clustering visual tokens to form semantic regions guided by text, whereas TCL \cite{tcl} leverages text supervision to generate dense pseudo-masks for cross-modal alignment. While these approaches offer an efficient alternative by alleviating reliance on dense masks, they remain fundamentally bottlenecked by the VLM's visual encoder.

\subsection{Self-Supervised Visual Backbones}\label{subsec:selfsup}

To compensate for the spatial limitations of VLM encoders, recent works have started utilizing Self-Supervised Vision Models (SSVMs). In particular, the DINO series \cite{dinov1, dinov2, dinov2_reg, dinov3} has shown that it can extract object-centric patch embeddings and maintain fine-grained spatial structures without relying on text supervision. Because of this, a common approach in OVSS is to directly align the spatial features of DINO with the semantic space of CLIP \cite{CLIP}.

The effectiveness of SSVMs in dense prediction mainly comes from their emergent object-centric properties. As observed in previous studies~\cite{simeoni2021localizing, wang2023tokencut}, the self-attention maps in these models naturally capture semantic boundaries and foreground objects without explicit labels. This localization capability allows SSVMs to act as a spatial anchor, which helps resolve the coarse resolution and grid artifacts commonly found in VLM outputs. Therefore, recent OVSS methods often adopt a dual-backbone strategy, using DINO for boundary refinement and CLIP~\cite{CLIP} for zero-shot classification~\cite{clip_dinoiser, lai2025exploring}. By combining global text-image alignment with local geometric priors, these frameworks produce much sharper mask boundaries than VLM-only baselines.

\subsection{Continuous and Flow-Based Alignment}\label{subsec:flows}
Neural ODEs~\cite{chen2018neuralode} introduced the idea of framing deep networks as continuous dynamic systems. Instead of a discrete sequence of layers, the network learns an ODE to transform the hidden features. This continuous-time formulation is now heavily used in generative modeling. Flow Matching~\cite{flowmatching, rectifiedflow} and Normalizing Flows~\cite{rezende2015variational}, for instance, rely on learning a vector field to map between data distributions.
%
In representation learning, ODE networks also guarantee invertibility and stability for robust feature extraction~\cite{behrmann2019invertible, dupont2019augmented}.
Furthermore, because many feature spaces are non-Euclidean, researchers proposed Riemannian and manifold ODEs~\cite{mathieu2020riemannian, lou2020neural}. By forcing the features to move strictly along geometric manifolds like a hypersphere, these models keep the original topological structure intact.
However, cross-modal alignment in OVSS still relies on discrete or static methods. Standard approaches simply use MLP layers or Optimal Transport~\cite{ovcoast_cvprw25}. These sudden mapping steps easily break the geometric curvature of the embeddings. While continuous flows naturally preserve this geometry, they are rarely used to align different modalities. 
Therefore, we introduce a manifold-constrained ODE flow that continuously maps the CLIP~\cite{CLIP} text space to the DINOv3~\cite{dinov3} visual space, achieving a smooth and geometry-aware alignment.

\section{Method}
In this work, we introduce an ODE-based alignment framework that continuously transforms text embeddings into the visual space.
Section~\ref{sec:3_preliminary} presents the fundamental formulation of the ODE used in our method and briefly describes the formulation of the DINOv3 vision encoder and the CLIP text encoder. 
Then, Section~\ref{sec:3_method} details our proposed \textbf{\sys} framework and its key components.

\subsection{Preliminaries}\label{sec:3_preliminary}
\paragraph{\textbf{Ordinary Differential Equations.}}
An ordinary differential equation (ODE) defines how a continuous variable evolves over time, governed by a vector field. Given a state $z_t \in \mathbb{R}^d$ and a time variable $t \in [0, 1]$, a neural ODE parameterizes the rate of change using a learnable neural network $v_\theta(z_t, t)$. The dynamics of $z_t$ are described by the initial value problem:
\begin{equation}\label{eq:ode}
    \frac{d z_t}{d t} = v_\theta(z_t, t), \quad z_{t=0} = z_0.
\end{equation}
Integrating this ODE from time $t=0$ to $t=1$ yields a continuous transformation trajectory from the initial state to the final state:
\begin{equation}
    z_1 = z_0 + \int_0^1 v_\theta(z_t, t) \, dt.
\end{equation}
In practice, this continuous integral is numerically approximated using discrete solvers such as the Euler method. For a step size $\Delta t = 1/N_{step}$, the state is iteratively updated as:
\begin{equation}\label{eq:ode_eulersolver}
    z_{t+\Delta t} = z_t + \Delta t \cdot v_\theta(z_t, t).
\end{equation}
This iterative update can be viewed as the \textit{continuous-depth limit of a residual network}, where each block performs an infinitesimal update along the time-dependent velocity field $v_\theta$. Unlike discrete, single-step projections (e.g., MLPs), the ODE formulation provides a smooth, invertible, and stable trajectory for modeling complex feature transformations.

\paragraph{\textbf{Task Formulation.}}
We build upon two complementary pre-trained models: DINOv3 for visual representation and CLIP for text embedding. 
Our goal is to bridge these two representations by learning a continuous transformation that evolves the CLIP semantic manifold toward DINOv3’s visual manifold for OVSS.
Given an input image $I \in \mathbb{R}^{3 \times H \times W}$, 
we denote the DINOv3 visual encoder as
$
f_{vis} : \mathbb{R}^{3 \times H \times W} \rightarrow \mathbb{R}^{(N+1) \times D}$.
The backbone encodes the image into $N$ non-overlapping patch tokens together with a learnable \texttt{[CLS]} token:
\begin{equation}
\left( \mathbf{z}^{img}_{cls}, \mathbf{Z}^{img} \right)=f_{vis}(I)  ,
\end{equation}
where $\mathbf{Z}^{img} \in \mathbb{R}^{N \times D}$ and 
$\mathbf{z}^{img}_{cls} \in \mathbb{R}^{D}$, 
with $D$ denoting the embedding dimension of DINOv3.
For a textual prompt $T$, we denote the CLIP text encoder as
$f_{text}$,
and the resulting semantic embedding is:
\begin{equation}
\mathbf{z}^{text} = f_{text}(T) \in \mathbb{R}^{D_{text}},
\end{equation}
where $D_{text}$ is the CLIP embedding dimension.
\begin{figure*}[t]
   \includegraphics[width=1.0\linewidth]{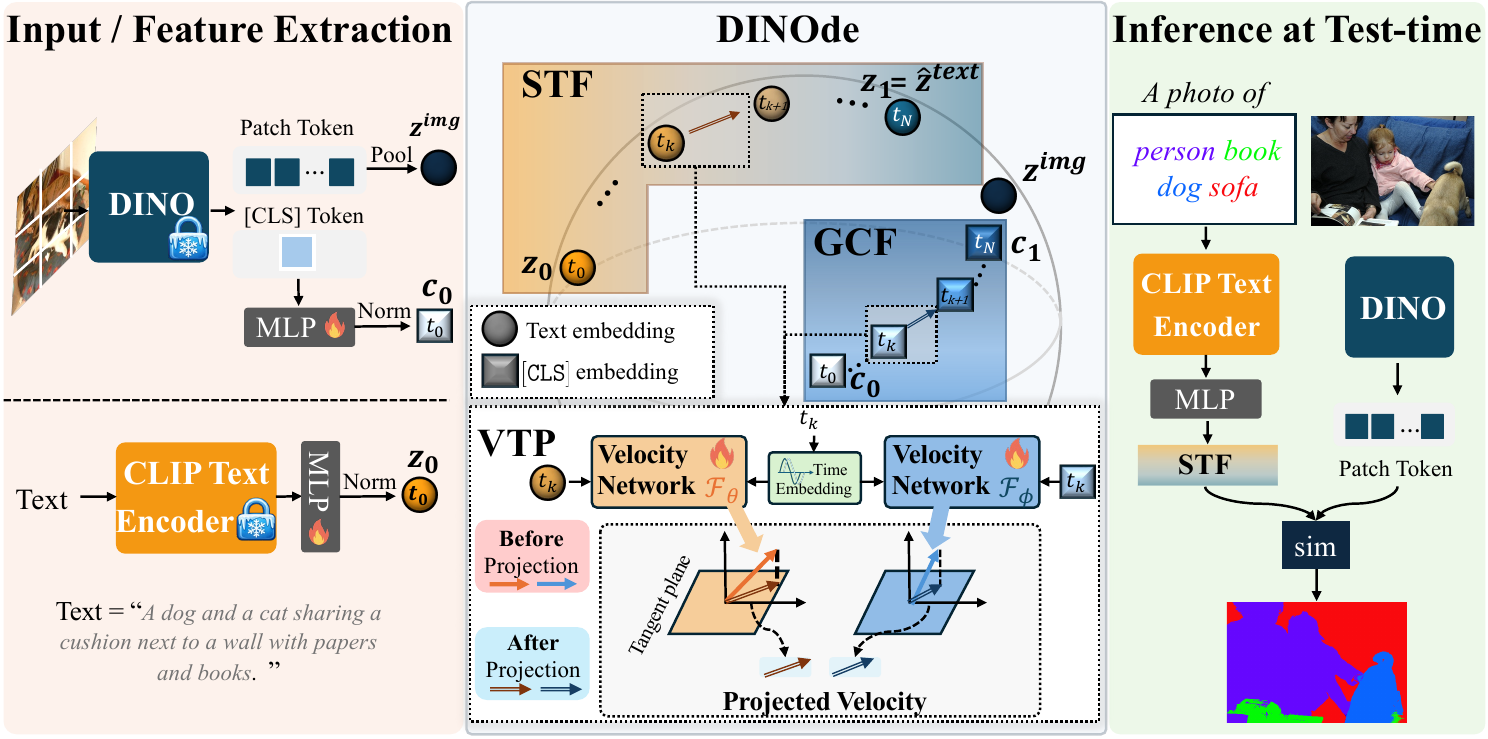}
    \caption{\textbf{Overview of the proposed \sys~framework.} Given an input image, a frozen DINOv3 encoder extracts patch tokens and a \texttt{[CLS]} token, while a CLIP text encoder produces a semantic text embedding. \textbf{Semantic Text Flow (STF)} continuously aligns the text embedding to the DINO visual manifold via an ODE-based transformation, while \textbf{Global Context Flow (GCF)} refines the global \texttt{[CLS]} representation. To preserve hyperspherical geometry, \textbf{Velocity Tangent Projection (VTP)} constrains the velocity to the tangent space. The aligned embeddings are used to compute patch–text similarity for open-vocabulary semantic segmentation.}
    \label{fig:method_framework}
\end{figure*}

\subsection{\sys}\label{sec:3_method}
In this work, we aim to exploit the potential of a self-supervised vision model for open-vocabulary semantic segmentation. 
We experimentally observe that the MLP-based method leads to suboptimal alignment, primarily because it employs a Euclidean shortcut that disregards the manifold of the feature space. 
To this end, we propose \textbf{\sys}, an ODE-based image-text alignment framework that continuously transforms CLIP text space toward the visual space of DINO. 
\sys~consists of two complementary components: 
(\emph{i}) \textbf{Semantic Text Flow (STF)} that gradually aligns the text manifold into the vision manifold, and 
(\emph{ii}) \textbf{Global Context Flow (GCF)} that also progressively refines the DINO's holistic \texttt{[CLS]} token, producing a globally aligned representation.

\paragraph{\textbf{Semantic Text Flow}.} 
As shown in Fig.~\ref{fig:method_framework}, to continuously align the text embedding with the visual manifold, we model the image-text alignment as an ODE as in Eq.~\ref{eq:ode}:
$
\frac{d\mathbf{z}_t}{dt}
= v_\theta(\mathbf{z}_t, t)$.
To specify the initial state and the visual supervision target,
we first map the text embedding to the DINO feature dimension via a learnable linear projection:
$\mathbf{z}_0 = \mathrm{Norm}(W \mathbf{z}^{text})$, where $W \in \mathbb{R}^{D \times D_{text}}$,
and use $\mathbf{z}_0$ as the initial condition. 
Then, we obtain a pooled visual embedding
\begin{equation}\label{eq:vis_emb}
\mathbf{z}^{img} = \mathrm{Norm}(\mathrm{Pool}(\mathbf{Z}^{img}))\in\mathbb{R}^{D},
\end{equation}
which serves as the target visual representation.
Here, $\mathrm{Norm}(\mathbf{x})=\mathbf{x}/\|\mathbf{x}\|_2$ denotes $\ell_2$ normalization.
For the $\mathrm{Pool}$ operation, we employ top-$K$ pooling as in~\cite{yoon2022adversarial}.
To effectively inject temporal information into the velocity network $\mathcal{F}_\theta$,
we realize time conditioning via a sinusoidal embedding $\gamma(t)$,
which is incorporated through feature-wise modulation, i.e.,
$v_\theta(\mathbf{z}, t)=\mathcal{F}_\theta(\mathbf{z};\gamma(t))$. The details about the velocity network are provided in the \textit{Supplementary Materials}.
To preserve the intrinsic geometry of the hyperspherical feature space,
we propose \textbf{Velocity Tangent Projection (VTP)} that constrains the velocity to the tangent space at $\mathbf{z}_t$:
\begin{equation}
\tilde{v}_\theta(\mathbf{z}_t, t)
=
v_\theta(\mathbf{z}_t, t)
-
\langle v_\theta(\mathbf{z}_t, t), \mathbf{z}_t \rangle \mathbf{z}_t.
\label{eq:tangent_proj}
\end{equation}
We then integrate the ODE forward from $t=0$ to $t=1$ to obtain the aligned embedding as in Eq.~\ref{eq:ode_eulersolver}:
\begin{equation}\label{eq:stf_output}
\hat{\mathbf{z}}^{text} = \mathbf{z}_{1} = \Phi_\theta^{text}(\mathbf{z}_0),
\end{equation}
where $\Phi_\theta^{text}$ denotes the numerical flow map obtained by integrating Eq.~\ref{eq:ode} with an explicit Euler solver (Eq.~\ref{eq:ode_eulersolver}),
i.e., for $N_{step}$ with $\Delta t = 1/N_{step}$,
\begin{equation}
\mathbf{z}_{t_{k+1}} = \mathrm{Norm}\big(\mathbf{z}_{t_k} + \Delta t\,\tilde{v}_\theta(\mathbf{z}_{t_k}, t_k)\big),
\quad t_k = k\Delta t.
\label{eq:stf_euler}
\end{equation}

\paragraph{\textbf{Global Context Flow.}}
In addition to text alignment, we introduce \textbf{Global Context Flow (GCF)} to refine the holistic image representation carried by the \texttt{[CLS]} token.
DINOv3 produces a global representation $\mathbf{z}^{img}_{cls}\in\mathbb{R}^{D}$ that summarizes the overall image context by aggregating information from all patch tokens.
While the patch tokens $\mathbf{Z}^{img}$ capture fine-grained, part-level details, the \texttt{[CLS]} token provides holistic information of the given image.
We leverage this property to align global semantics between the visual and textual spaces.
Given the raw \texttt{[CLS]} token $\mathbf{z}^{img}_{cls}$ from the frozen DINOv3 backbone, we first apply a lightweight projection head and $\ell_2$ normalization to obtain an initial state on the hypersphere:
\begin{equation}
\mathbf{c}_0 = \mathrm{Norm}\big(W_{cls}\mathbf{z}^{img}_{cls}\big) \in \mathbb{R}^{D},
\label{eq:cls_init}
\end{equation}
where $W_{cls}\in \mathbb{R}^{D\times D}$.
We then evolve $\mathbf{c}_0$ via an ODE as in STF:
\begin{equation}
\frac{d\mathbf{c}_t}{dt} = u_\phi(\mathbf{c}_t, t),
\label{eq:cls_ode}
\end{equation}
where $u_\phi$ is a learnable velocity network.
Similar to STF, time conditioning is realized using a sinusoidal embedding $\gamma(t)$ injected via feature-wise modulation.
Here, we also apply \textbf{VTP} to preserve the hyperspherical geometry:
\begin{equation}
\tilde{u}_\phi(\mathbf{c}_t, t)
=
u_\phi(\mathbf{c}_t, t)
-
\langle u_\phi(\mathbf{c}_t, t), \mathbf{c}_t \rangle \mathbf{c}_t.
\label{eq:cls_vtp}
\end{equation}
Finally, the globally refined representation is obtained by forward integration:
\begin{equation}
\hat{\mathbf{z}}^{img}_{cls} =\mathbf{c}_{1}= \Psi_\phi(\mathbf{c}_0),
\label{eq:gcf_output}
\end{equation}
where $\Psi_\phi$ denotes the numerical flow map induced by Eq.~\ref{eq:cls_ode}.
Again, we use Euler solver with $N_{step}$ and step size $\Delta t = 1/N_{step}$ as follows:
\begin{equation}
\mathbf{c}_{t_{k+1}} = \mathrm{Norm}\big(\mathbf{c}_{t_{k}} + \Delta t\, \tilde{u}_\phi(\mathbf{c}_{t_k}, t_k)\big), \quad t_k = k\Delta t.
\label{eq:cls_euler}
\end{equation}

\paragraph{\textbf{Loss formulation.}}
In \sys, we adopt a CLIP-style symmetric contrastive objective to train the velocity networks $(v_\theta, u_\phi)$. Here, motivated by the prior work~\cite{dinotxt}, 
we construct a global image representation by concatenating the pooled patch descriptor $\mathbf{z}^{img}$ and the refined \texttt{CLS} token $\hat{\mathbf{z}}^{img}_{cls}$ obtained via GCF:
\begin{equation}
\mathbf{z}^{img}_{||} =
\Big[
\mathbf{z}^{img}\,;\,
\hat{\mathbf{z}}^{img}_{cls}
\Big]
\in \mathbb{R}^{2D}.
\label{eq:global_img_feat}
\end{equation}
For each image-caption pair, we encode the caption with CLIP and produce a vision-aligned text feature
$\hat{\mathbf{z}}^{text} \in \mathbb{R}^{D}$ obtained with STF (Eq.~\ref{eq:stf_output}).
To match dimensionality with $\mathbf{z}^{img}_{||}$, we repeat the $\hat{\mathbf{z}}^{text}$ to form $\mathbf{z}^{text}_{||}\in \mathbb{R}^{2D}$.
Given a minibatch of $B$ image-caption pairs, we compute the similarity matrix
\begin{equation}
\mathbf{S}_{ij} = \frac{1}{\tau}\,(\mathbf{z}^{img}_{||,i})^\top \mathbf{z}^{text}_{||,j},
\label{eq:sim_matrix}
\end{equation}
where $\tau$ is a temperature.
The symmetric contrastive loss is then defined as
\begin{equation}
\mathcal{L}_{NCE}
=
\frac{1}{2}\Big(
\mathrm{CE}(\mathbf{S}, \mathbf{y})
+
\mathrm{CE}(\mathbf{S}^\top, \mathbf{y})
\Big),
\qquad
\mathbf{y}=[1,\ldots,B],
\label{eq:nce}
\end{equation}
where $\mathrm{CE}$ and $\mathbf{y}$ are cross-entropy loss and matching indices, respectively.

\paragraph{\textbf{Inference.}}
As illustrated in Fig.~\ref{fig:method_framework}, the inference stage of \sys~utilizes the learned STF to generate semantic anchors for arbitrary categories and performs pixel-level segmentation by measuring their similarity with DINOv3 patch tokens. Specifically, for a given set of $M$ candidate categories $\{c_m\}_{m=1}^M$, we first extract their textual embeddings using a frozen CLIP text encoder and map them to the source state $\mathbf{z}_0 \in \mathbb{R}^{D\times M}$ via the learnable projection $W$. We then perform forward integration from $t=0$ to $t=1$ by utilizing the learned velocity network $v_\theta$ as described in Eq.~\ref{eq:stf_euler}. The resulting target state, $\hat{\mathbf{z}}^{text} = \mathbf{z}_1 \in \mathbb{R}^{D \times M}$, serves as a set of semantic anchors optimally aligned with the visual manifold of DINO. Finally, the dense segmentation map is generated by computing the cosine similarity between the DINO patch tokens and these aligned semantic anchors $\hat{\mathbf{z}}^{text}$.

\section{Experiments}

\subsection{Experimental Settings}

\paragraph{\textbf{Datasets.}} 
To evaluate the robustness and generalizability of our proposed method, we conduct extensive experiments on eight common OVSS benchmarks. Following the categorization in prior literature~\cite{tcl, freeda, clip_dinoiser, talk2dino}, these datasets are divided into two groups based on the inclusion of a background (bg) class.
The first group consists of three benchmarks with a background class: \textbf{Pascal VOC 2012 (V21)}~\cite{pascalvoc}, \textbf{Pascal Context (C60)}~\cite{pascalcontext}, and \textbf{COCO Object (Object)}~\cite{cocostuff}, containing 21, 60, and 81 classes, respectively.
The second group includes five benchmarks without a background class. Specifically, we evaluate on \textbf{COCO Stuff (Stuff)}~\cite{cocostuff}, \textbf{Cityscapes (City)}~\cite{cityscape}, and \textbf{ADE20K (ADE)}~\cite{ade20k_1,ade20k_2}, which comprise 171, 19, and 150 classes, respectively. Furthermore, we report performance on \textbf{Pascal VOC 2012 (V20)}~\cite{pascalvoc} and \textbf{Pascal Context (C59)}~\cite{pascalcontext}, both of which exclude the background class to provide a comprehensive comparison.

\paragraph{\textbf{Evaluation Protocol and Metrics.}} 
We follow the evaluation protocols established in previous works~\cite{groupvit22,tcl,clip_dinoiser,talk2dino}. During inference, for a fair comparison with the state-of-the-art work~\cite{talk2dino}, input images are resized to a shorter-side length of 448 pixels and processed via sliding window inference with a $448 \times 448$ resolution and a 224-pixel stride.
Following prior works~\cite{talk2dino,sun2025cliper}, during inference, the textual embeddings are generated by averaging the encodings of standard prompt templates (\textit{e.g.}, ``\textit{a photo of a \{class\}}'').
The primary evaluation metric for every experiment is the mean-Intersection-over-Union (mIoU).

\paragraph{\textbf{Implementation Details.}} 
Throughout the experiments, we employ DINOv3 ViT-L/16 as our visual backbone and CLIP ViT-L/14 as the textual encoder.
In contrast to conventional weakly-supervised OVSS methods~\cite{groupvit22,luo2023segclip,tcl,ovsegmentor,zeroseg} that rely on large-scale image-caption datasets such as CC3M~\cite{cc3m} ($\sim$3.3M) or CC12M~\cite{cc12m} ($\sim$12.4M), our DINO and ODE-based approach achieves efficient alignment with significantly less data, specifically training on the COCO 2017 Caption~\cite{coco,coco_caption} \textit{train} set ($\sim$118k images).
We use the AdamW optimizer with a learning rate of $1 \times 10^{-4}$ and a weight decay of $0.01$. The model is trained for 20 epochs with a batch size of 256 on a \underline{single} NVIDIA RTX 3090 GPU.
Since we utilize frozen DINOv3 and CLIP encoders, we can leverage pre-cached embeddings, and the aligned text anchors are cached once per class set, allowing further computational efficiency.
The entire training process, including the initial caching phase, is completed within 4 hours of wall-clock time.
Consistent with~\cite{tcl, talk2dino}, we apply Pixel-Adaptive Mask Refinement (PAMR)~\cite{pamr_postprocessing} as a post-processing step to enhance the boundary adherence of the final semantic maps.
The hyperparameters for our method are as follows: the number of ODE steps $N_{step}=10$, $K=20$ for top-$K$ pooling, the temperature $\tau=0.07$ for NCE loss.

\subsection{Discussions}

\begin{figure}[t]
  \centering
   \includegraphics[width=1.0\linewidth]{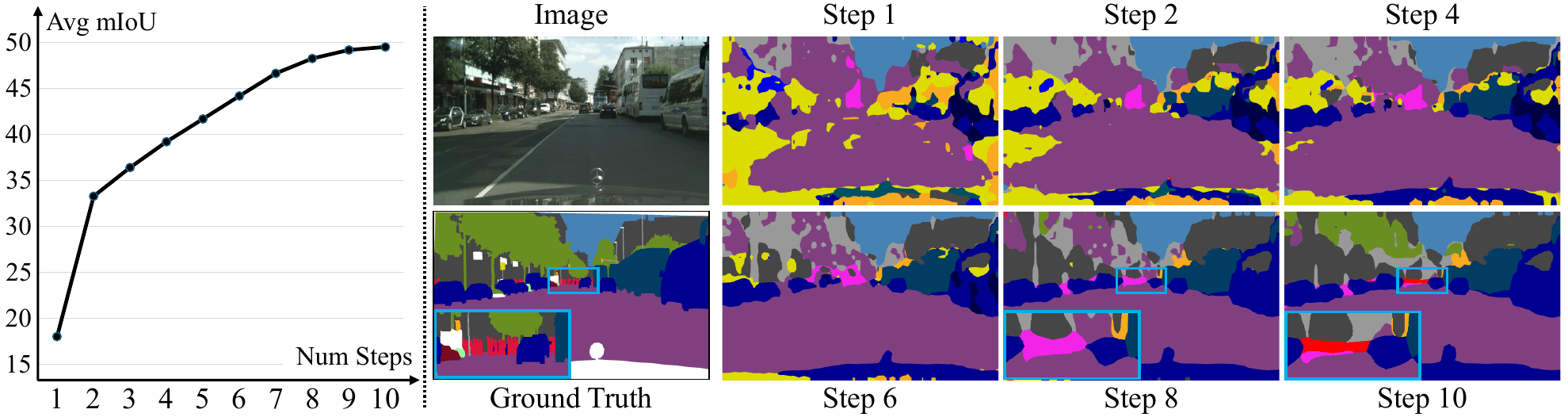}
   \caption{
    \textbf{Analysis of Semantic Text Flow (STF) with varying numbers of ODE steps}. Segmentation maps are obtained using the aligned text embedding $\mathbf{z}_{t_k}$ at each step $k \in \{1,2,\ldots,10\}$. 
   (Left) Average mIoU across eight benchmarks, showing the progressive performance improvement during the manifold transition. (Right) Qualitative results on Cityscapes across steps.}
   \label{fig:per_step}
\end{figure}

\paragraph{\textbf{Analysis of Progressive Manifold Transitions.}}
To verify whether STF effectively learns the progressive manifold transition via ODE as intended, we perform quantitative and qualitative comparisons across varying numbers of ODE integration steps.
The average mIoU across eight benchmarks for each step is illustrated in the graph of Fig.~\ref{fig:per_step} (left). Notably, the performance exhibits a rapid increase from steps 1 to 8. This trend demonstrates that the velocity network is effectively trained to transport CLIP text embeddings toward the DINO visual manifold in a progressive and stable manner.
Between steps 8 and 10, the mIoU curve shows a more gradual yet steady improvement. This suggests that once the embeddings reach the vicinity of the DINO manifold, the STF module performs fine-grained adjustments to converge toward the optimal semantic coordinates within the visual space.
The qualitative results on the Cityscapes dataset, shown in Fig.~\ref{fig:per_step}, consistently align with these quantitative observations. In the first 8 steps, we observe rapid improvements in global structure and semantic consistency. 
In the subsequent steps, the results exhibit further improvements in fine details.
In conclusion, these results demonstrate that our STF successfully learns an optimal manifold transition trajectory, effectively bridging the modality gap between CLIP and DINO.

\begin{table*}[t]
\centering
\caption{Component analysis of the proposed method.}
\label{tab:component_abl}
\resizebox{0.95\textwidth}{!}{
\begin{tabular}{c|ccc|cccccccc|c}
\toprule
& \textbf{STF} & \textbf{VTP} & \textbf{GCF} & \textbf{V20} & \textbf{C59} & \textbf{Stuff} & \textbf{City} & \textbf{ADE} & \textbf{V21} & \textbf{C60} & \textbf{Object} & \textbf{Avg} \\
\midrule
(a) &            &            &            & 88.4 & 43.2 & 29.7 & 42.5 & 23.2 & 68.0 & 39.6 & 46.7 & 47.7 \\
(b) & \checkmark &            &            & 88.4 & 43.0 & 31.2 & 44.7 & 25.2 & 66.6 & 39.5 & 47.1 & 48.2 \\
(c) & \checkmark & \checkmark &            & 88.5 & 43.4 & 31.4 & 45.4 & 25.5 & 67.3 & 39.6 & 47.2 & 48.5 \\
(d) & \checkmark &            & \checkmark & 90.2 & 44.3 & 31.3 & 45.6 & 24.8 & 69.5 & 40.3 & 47.2 & 49.2 \\
(e) & \checkmark & \checkmark & \checkmark & 91.1 & 44.5 & 31.6 & 45.4 & 25.4 & 69.2 & 40.7 & 48.0 & \textbf{49.5} \\

\bottomrule
\end{tabular}
}
\end{table*}

\paragraph{\textbf{Component Analysis.}}
Table~\ref{tab:component_abl} presents an ablation study evaluating the individual contributions of our proposed Semantic Text Flow (STF) and Global Context Flow (GCF), as well as the impact of Velocity Tangent Projection (VTP).
In this table, result (a) is the baseline where text-to-visual alignment is performed using a standard MLP with a parameter capacity \underline{comparable} to that of the STF Velocity Network.
A comparison between (a) and (b)–(e) demonstrates that the ODE-based progressive transition via STF achieves superior performance compared to simple static mapping. Notably, the steady improvement in average mIoU from (a) to (c) and (e) suggests that STF successfully learns an optimal transition trajectory for transferring CLIP text embeddings to the DINO visual manifold, surpassing the capabilities of mere feature alignment.
To verify the role of VTP in preserving the intrinsic geometry of the hyper-spherical feature space, we compare (b) with (c) and (d) with (e); in both cases, VTP leads to meaningful performance improvements.
This demonstrates that projecting velocity vectors onto the tangent space, thereby maintaining the feature norm during transition, plays an important role in ensuring the stability and accuracy of the progressive manifold transfer.
Finally, we analyze the effect of GCF, which leverages the DINO \texttt{[CLS]} token to incorporate global contextual information. GCF ensures that the transition of text embeddings (STF) is not biased toward local visual features but instead accounts for the broader global context. Comparisons between (b) and (d), as well as (c) and (e), demonstrate that GCF yields consistent performance improvements. 
These results indicate that STF and GCF complementarily optimize the continuous alignment between manifolds.

\begin{table*}[t]
\centering
\caption{Ablation study on the number of ODE steps ($N_{step}$).}
\label{tab:step_num_abl}
\resizebox{0.90\textwidth}{!}{
\begin{tabular}{c|c|cccccccc|c}
\toprule
\textbf{Align Method} & \textbf{$N_{step}$} & \textbf{V20} & \textbf{C59} & \textbf{Stuff} & \textbf{City} & \textbf{ADE} & \textbf{V21} & \textbf{C60} & \textbf{Object} & \textbf{Avg} \\
\midrule
MLP-based                  & -   & 88.4 & 43.2 & 29.7 & 42.5 & 23.2 & 68.0 & 39.6 & 46.7 & 47.7 \\ \hline
\multirow{3}{*}{ODE-based} & 5   & 89.7 & 44.4 & 31.7 & 43.5 & 25.2 & 66.7 & 40.4 & 48.0 & 48.7 \\
                           & 10  & 91.1 & 44.5 & 31.6 & 45.4 & 25.4 & 69.2 & 40.7 & 48.0 & 49.5 \\
                           & 50  & 91.0 & 44.1 & 31.9 & 46.1 & 25.4 & 69.1 & 40.2 & 48.5 & 49.6 \\

\bottomrule
\end{tabular}
}
\end{table*}

\paragraph{\textbf{Ablation on Number of Steps.}}
Table~\ref{tab:step_num_abl} analyzes the impact of the number of ODE steps ($N_{step}$) used during both the training and inference of the STF and GCF modules. 
We observe that even with $N_{step}=5$, our method outperforms the MLP-based discrete mapping, demonstrating the effectiveness of ODE-based transitions for cross-modal alignment. 
The improvement attained at $N_{step}=10$ further suggests that a sufficient level of granularity enables the model to more successfully navigate the complex manifold gap between the disparate domains.
For $N_{step} \ge 10$, the average mIoU exhibits a subtle upward trend as the number of steps increases. This suggests that a higher number of steps facilitates a smoother transition trajectory, assisting the model in converging more closely to the optimal position on the target manifold. However, we also observe marginal performance drops in certain benchmarks. This can be attributed to the accumulation of minute errors during the numerical integration process over an excessive number of steps, which potentially causes the final embedding to `over-shoot' its optimal target within the visual manifold. 
Considering the trade-off between performance gains and computational overhead, we select $N_{step}=10$ as the optimal balance for our framework.

\paragraph{\textbf{Generalization to Diverse Backbones.}}

\begin{table*}[t]
\centering
\caption{Generalization capability across various visual and textual backbones. We compare our proposed alignment (Ours) against the discrete projection baseline (MLP).}
\label{tab:backbone_generalization}
\resizebox{\textwidth}{!}{
\begin{tabular}{cc|c|cccccccc|c}
\toprule
\textbf{DINOv3(Visual)} & \textbf{CLIP(Textual)} & \textbf{Alignment} & \textbf{V20} & \textbf{C59} & \textbf{Stuff} & \textbf{City} & \textbf{ADE} & \textbf{V21} & \textbf{C60} & \textbf{Object} & \textbf{Avg} \\
\midrule
\multirow{2}{*}{ViT-L} & \multirow{2}{*}{ViT-L} & MLP & 88.4 & 43.2 & 29.7 & 42.5 & 23.2 & 68.0 & 39.6 & 46.7 & 47.7 \\
                               &                    & \textbf{Ours} & 91.1 & 44.5 & 31.6 & 45.4 & 25.4 & 69.2 & 40.7 & 48.0 & \textbf{49.5} \\
\midrule
\multirow{2}{*}{ViT-B} & \multirow{2}{*}{ViT-L} & MLP & 87.3 & 39.2 & 27.7 & 38.1 & 20.7 & 62.4 & 36.1 & 41.8 & 44.2 \\
 & & \textbf{Ours} & 89.2 & 41.6 & 29.3 & 37.7 & 22.4 & 66.2 & 38.4 & 43.8 & \textbf{46.1} \\
\midrule
\multirow{2}{*}{ViT-S} & \multirow{2}{*}{ViT-L} & MLP & 80.3 & 34.8 & 19.9 & 25.8 & 15.7 & 54.2 & 32.7 & 31.6 & 36.9 \\
                               &                    & \textbf{Ours} & 84.2 & 36.1 & 22.6 & 29.4 & 16.3 & 55.6 & 33.6 & 32.9 & \textbf{38.8} \\
\midrule

\multirow{2}{*}{ViT-L} & \multirow{2}{*}{ViT-B} & MLP & 87.7 & 43.1 & 30.6 & 43.8 & 24.3 & 67.3 & 39.5 & 46.3 & 47.8 \\
& & \textbf{Ours} & 90.8 & 42.8 & 30.5 & 42.8 & 25.0 & 69.8 & 39.7 & 48.1 & \textbf{48.7} \\
\bottomrule
\end{tabular}
}
\end{table*}

To demonstrate the generalization capability of the \sys, we conduct experiments with various visual and textual backbones, as shown in Table~\ref{tab:backbone_generalization}.
As the DINOv3 visual backbone scales down to ViT-B and ViT-S, our continuous flow-based alignment method consistently outperforms the baseline, MLP-based alignment. This suggests that, regardless of the granularity of visual representation, ODE-based trajectory modeling is inherently more effective at bridging the manifold gap between modalities than discrete projection methods.
Furthermore, \sys~surpasses the MLP-based baseline even when the CLIP textual backbone is transitioned to ViT-B. This demonstrates that our method effectively guides the transition trajectory toward the target DINO visual manifold, even when the underlying composition of the text embedding space is altered. 
These results demonstrate that the \sys~possesses high generalization capability, flexibly accommodating diverse visual and textual manifolds without relying on a specific backbone.

\begin{table}[t]
\centering
\caption{Geometric diagnostics of manifold preservation. STF consistently
outperforms the MLP baseline across all criteria.}
\label{tab:supp_geometry}
\setlength{\tabcolsep}{8pt}
\begin{tabular}{lcc}
\toprule
\textbf{Geometric Criterion} & \textbf{MLP} & \textbf{STF(Ours)}   \\
\midrule
(i) \textit{Neighborhood preservation}       & 0.575 & \textbf{0.601}  \\
(ii) \textit{Angular / geodesic consistency} & 0.693 & \textbf{0.712}  \\
(iii) \textit{Class-structure preservation}  & 0.843 & \textbf{0.871}  \\
(iv) \textit{Alignment-space compactness}    & 0.412 & \textbf{0.423}  \\
\bottomrule
\end{tabular}
\end{table}

\paragraph{\textbf{Geometric Diagnostics of Manifold Preservation.}}
To directly verify that STF preserves the intrinsic geometry of the feature space, we report four geometric diagnostics in Table~\ref{tab:supp_geometry}. 
We measure (i) the \textit{top-10 nearest-neighbor overlap} between the original CLIP class space and the aligned space, 
(ii) the \textit{local Pearson correlation of geodesic-distance vectors} between neighboring classes, 
(iii) linear \textit{CKA between class-level Gram matrices} to assess whether the overall semantic structure is preserved, and 
(iv) the \textit{mean cosine similarity} between aligned text features and mask-aware DINO patch prototypes of the corresponding classes. 
Across all four criteria, STF consistently surpasses the MLP baseline, indicating that the continuous ODE-based transition better retains neighborhood structure, geodesic consistency, and class-level semantic organization during cross-modal alignment.

\begin{table*}[t]
\centering
\caption{Quantitative comparison with state-of-the-art methods. \textbf{Bold} and \underline{Underline} denote the best and second-best results, respectively.}
\label{tab:quan_main}
\resizebox{\textwidth}{!}{
\begin{tabular}{llccccccccc}
\toprule
\textbf{Model} & \textbf{Visual Encoder} & \textbf{V20} & \textbf{C59} & \textbf{Stuff} & \textbf{City} & \textbf{ADE} & \textbf{V21} & \textbf{C60} & \textbf{Object} & \textbf{Avg} \\
\midrule
\multicolumn{11}{l}{\textit{without Mask Refinement}} \\
MaskCLIP~\cite{maskclip22}$_{ECCV22}$ & CLIP             & 29.4 & 12.4 & 8.8  & 11.5 & 7.2  & 23.3 & 11.7 & 7.2  & 13.9 \\
SCLIP~\cite{sclip}$_{ECCV24}$         & CLIP             & 70.6 & 25.2 & 17.6 & 21.3 & 10.9 & 44.0 & 22.3 & 26.9 & 29.9 \\
ClearCLIP~\cite{clearclip}$_{ECCV24}$ & CLIP             & 80.0 & 29.6 & 19.9 & 27.9 & 15.0 & - & - & - & - \\
NACLIP~\cite{naclip}$_{WACV25}$       & CLIP             & 78.7 & 32.1 & 21.4 & 31.4 & 17.3 & 52.2 & 28.7 & 29.9 & 36.5 \\
dino.txt~\cite{dinotxt}$_{CVPR25}$    & DINOv2(reg)      & 62.1 & 30.9 & 20.9 & 32.1 & 20.6 & - & - & - & - \\
FreeDA~\cite{freeda}$_{CVPR24}$       & DINOv2           & 71.8 & 35.4 & 24.2 & 32.3 & 19.4 & 44.9 & 31.1 & 24.6 & 35.5 \\
FreeDA~\cite{freeda}$_{CVPR24}$       & CLIP+DINOv2      & 85.7 & 39.7 & 26.3 & 33.6 & 21.4 & 44.1 & 34.8 & 33.9 & 39.9 \\
ProxyCLIP~\cite{proxyclip}$_{ECCV24}$ & CLIP+DINOv2(reg) & 85.2 & 36.2 & 24.6 & 35.2 & 21.6 & 56.6 & 33.0 & 36.7 & 41.1 \\
ProxyCLIP~\cite{proxyclip}$_{ECCV24}$ & CLIP+DINO        & 83.2 & 37.7 & 25.6 & 40.1 & 22.6 & 60.6 & 34.5 & 39.2 & 43.0 \\
CLIPer~\cite{sun2025cliper}$_{ICCV25}$ & CLIP            & 88.2 & 39.8 & 25.8 &   -  & 21.8 & 61.2 & 34.3 & 39.6 &   -  \\
Talk2DINO~\cite{talk2dino}$_{ICCV25}$ & DINOv2(reg)      & 87.1 & 39.1 & 27.0 & 35.8 & 21.1 & 60.1 & 34.2 & 37.6 & 42.8 \\
Talk2DINO*~\cite{talk2dino}$_{ICCV25}$ & DINOv3          & \underline{87.7} & \underline{43.0} & \textbf{32.6} & \underline{40.8} & \underline{24.2} & \underline{65.9} & \underline{37.9} & \textbf{48.6} & \underline{47.6} \\
Ours                                  & DINOv3           & \textbf{91.1} & \textbf{44.5} & \underline{31.6} & \textbf{45.4} & \textbf{25.4} & \textbf{69.2} & \textbf{40.7} & \underline{48.0} & \textbf{49.5} \\
\midrule
\multicolumn{11}{l}{\textit{with Mask Refinement}} \\
SCLIP~\cite{sclip}$_{ECCV24}$         & CLIP             & 76.3 & 27.4 & 18.7 & 23.9 & 11.8 & 47.8 & 23.8 & 26.9 & 32.1 \\
NACLIP~\cite{naclip}$_{WACV25}$       & CLIP             & 84.5 & 36.4 & 24.6 & 37.1 & 19.6 & 57.9 & 36.4 & 34.6 & 41.4 \\
FreeDA~\cite{freeda}$_{CVPR24}$       & DINOv2           & 75.2 & 39.0 & 27.0 & 33.1 & 21.3 & 45.3 & 34.3 & 26.7 & 37.7 \\
FreeDA~\cite{freeda}$_{CVPR24}$       & CLIP+DINOv2      & 87.1 & 42.4 & 28.1 & 33.8 & 22.6 & 55.4 & 37.1 & 36.1 & 42.8 \\
ProxyCLIP~\cite{proxyclip}$_{ECCV24}$ & CLIP+DINOv2(reg) & 85.8 & 37.6 & 25.6 & 37.5 & 22.5 & 59.4 & 34.6 & 39.0 & 42.8 \\
ProxyCLIP~\cite{proxyclip}$_{ECCV24}$ & CLIP+DINO        & 83.2 & 38.0 & 26.2 & \underline{41.0} & 22.6 & 60.7 & 34.7 & 39.4 & 43.2 \\
CLIPer~\cite{sun2025cliper}$_{ICCV25}$ & CLIP            & \underline{90.0} & 43.6 & 28.7 &   -  & 24.4 & \textbf{69.8} & 38.0 & 43.3 &   -  \\
Talk2DINO~\cite{talk2dino}$_{ICCV25}$ & DINOv2(reg)      & 89.8 & 42.7 & 29.6 & 38.4 & 22.9 & 66.1 & 37.3 & 42.3 & 46.1 \\
Talk2DINO*~\cite{talk2dino}$_{ICCV25}$ & DINOv3          & 88.2 & \underline{44.3} & \textbf{33.8} & 40.5 & \underline{24.8} & \underline{67.0} & \underline{39.2} & \textbf{50.4} & \underline{48.5} \\
Ours                                  & DINOv3           & \textbf{91.6} & \textbf{45.3} & \underline{32.1} & \textbf{46.2} & \textbf{25.9} & \textbf{69.8} & \textbf{41.1} & \underline{48.4} & \textbf{50.1} \\
\bottomrule
\end{tabular}
}
\end{table*}

\begin{figure}[t]
  \centering
  \includegraphics[width=1.0\linewidth]{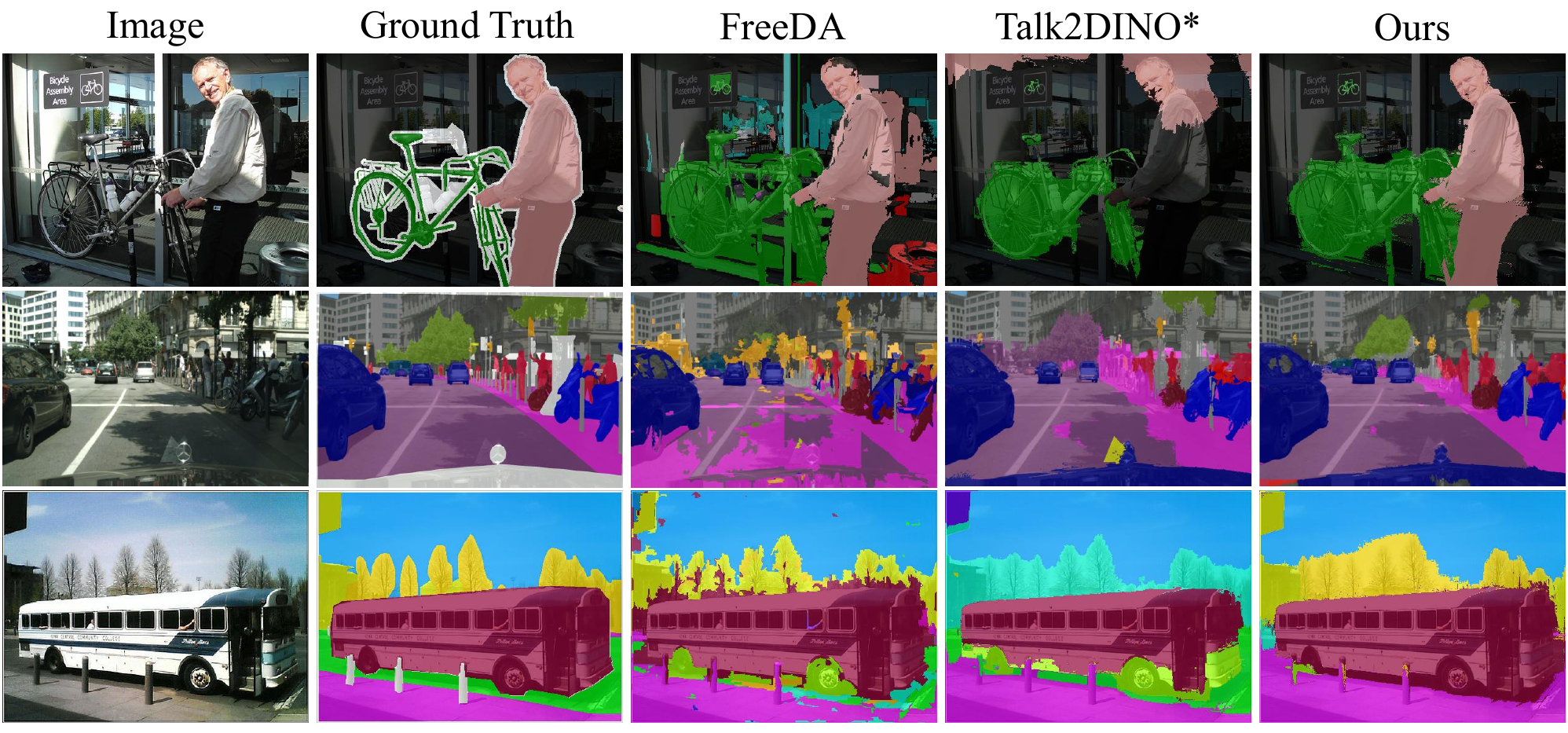}
   \caption{Qualitative comparison with the state-of-the-art methods. The result of Talk2DINO* is obtained using the DINOv3 backbone for a fair comparison.}
   \label{fig:qual_main}
\end{figure}

\paragraph{\textbf{Comparison to state-of-the-art.}}
In Table~\ref{tab:quan_main}, we present a quantitative comparison between our proposed \sys~and state-of-the-art OVSS methods across eight benchmarks.
Baseline methods are categorized into training-free OVSS approaches~\cite{maskclip22, shin2022reco, bousselham2024grounding, sclip, clearclip, sun2024clip, luo2024emergent, naclip, freeda,proxyclip,sun2025cliper} and weakly-supervised OVSS methods~\cite{groupvit22, luo2023segclip, tcl, ovseg23, mukhoti2023open, yi2023simple, cai2023mixreorg,dinotxt,talk2dino} that leverage image-caption pairs. 
We conduct comprehensive comparisons with representative baselines and recent state-of-the-art approaches, particularly dino.txt~\cite{dinotxt} and Talk2DINO~\cite{talk2dino}, which also utilize image-caption pairs for training.
To ensure a fair comparison, the results for Talk2DINO* in Table~\ref{tab:quan_main} and Fig.~\ref{fig:qual_main} were obtained using officially released checkpoints and source code integrated with a DINOv3 backbone.
Specifically for benchmarks with a background, we conducted an exhaustive search for the optimal background threshold to ensure peak performance of Talk2DINO*.

Our method outperforms existing state-of-the-art approaches on most benchmarks (6 out of 8), both with and without mask refinement. 
Notably, compared with Talk2DINO*, which employs the same DINOv3 visual encoder backbone, our approach achieves an average mIoU of 49.5\% before mask refinement, yielding a 1.9\%p improvement. 
Upon incorporating the final mask refinement step, our framework reaches a peak average mIoU of 50.1\%, further consolidating its superior performance across most of the benchmarks.
This result suggests that the gains are not merely due to the superior visual representations of the backbone.
The superiority of our approach is further supported by the qualitative comparisons presented in Fig.~\ref{fig:qual_main}, which illustrates the results across three distinct benchmarks: Pascal VOC (V21), Cityscapes (City), and Pascal Context (C59). 
Notably, in the Cityscapes examples (Fig.~\ref{fig:qual_main} middle row), our method successfully captures small \textit{Bus} and \textit{vegetation} classes that the compared baselines fail to identify.
Additional qualitative comparison results are provided in the \textit{Supplementary Materials.}
Our results exhibit noticeable improvements in fine details compared to other SOTA methods, demonstrating that the ODE-based transition effectively achieves optimal textual-visual manifold alignment.

\section{Conclusion}
In this work, we propose \sys, an ODE-based alignment framework that bridges the semantic gap between self-supervised visual representations and text representations for open-vocabulary semantic segmentation. Existing MLP-based projection methods approximate the complex relationship between heterogeneous representation manifolds through a single-step mapping, which often fails to preserve the intrinsic geometry of the feature space. To address this limitation, in \sys, we propose Semantic Text Flow (STF) that gradually evolves text embeddings toward the DINO feature space along an ODE trajectory, enabling smooth cross-modal alignment. In addition, Global Context Flow (GCF) progressively refines the global image representation carried by the \texttt{[CLS]} token to better capture holistic visual semantics. To further preserve the hyperspherical geometry of the feature space, we introduce Velocity Tangent Projection (VTP), which constrains the velocity field to the tangent space and enables geometry-preserving alignment. Extensive experiments demonstrate that the proposed \sys~effectively improves cross-modal representation learning and consistently enhances performance on OVSS benchmarks. 
These results highlight the potential of flow-based alignment for bridging vision and language representations. As future work, since modern multimodal large language models map frozen visual features into the language space via a single MLP projector, which is analogous to the discrete baseline we revisit, we expect continuous ODE-based alignment to be a promising direction for improving fine-grained visual grounding in MLLMs.
\section*{Acknowledgements}
This research was supported by the National Research Foundation (NRF) funded by the Korean government (MSIT) (No. RS-2026-25561904) and by the National Research Foundation of Korea(NRF) grant funded by the Korea government(MSIT) (RS-2026-25473963).
\bibliographystyle{splncs04}
\bibliography{main}
\clearpage
\setcounter{section}{0}
\renewcommand{\thesection}{S\arabic{section}}

\renewcommand{\theHsection}{supp.\arabic{section}}

\begin{center}
    {\Large\bfseries \textit{Supplementary Materials}}
\end{center}

\section{\textit{Additional Details}}
\subsection{\textit{Implementation Details}}
We strictly adhere to the experimental protocols established in prior state-of-the-art works~\cite{sun2025cliper, talk2dino} regarding prompt templates, background handling, and mask refinement(PAMR) settings to ensure a fair and consistent comparison. 

\noindent{\textit{\textbf{Background Handling.}}} The strategy for background management varies depending on the benchmark characteristics. For benchmarks without a dedicated background class (V20, C59, Stuff, City, and ADE), we restrict the model to predict only foreground classes, following Talk2DINO~\cite{talk2dino}. In these instances, the background annotations provided in the ground truth (GT) for Pascal VOC, Context, and ADE20K are treated as ignored regions.
For benchmarks that include a background category (V21, C60, and Object), we follow CLIPer~\cite{sun2025cliper}. We incorporate the background into the prediction space and utilize background thresholding to reassign regions with confidence scores below a certain threshold to the background class. Following the strict implementation of CLIPer, we employ the benchmark-specific background classes listed in Table~\ref{tb:bg_class}. When a benchmark contains multiple background categories, the final background probability is computed as the maximum value among the predicted background class probabilities.

\noindent{\textit{\textbf{Prompt Templates.}}}
For generating text embeddings via the CLIP text encoder, we adopt the specific prompt templates utilized in CLIPer~\cite{sun2025cliper}.

\noindent\textit{\textbf{Mask Refinement.}} For mask refinement, we adopt Pixel-Adaptive Mask Refinement (PAMR) following Talk2DINO~\cite{talk2dino}. Specifically, we apply local neighborhood kernels with dilation rates of $\{1, 2, 4, 8, 12, 24\}$ and run 10 refinement iterations to refine the predicted mask.

\subsection{\textit{Architectural Details}}

\begin{figure}[t]
  \centering
   \includegraphics[width=0.7\linewidth]{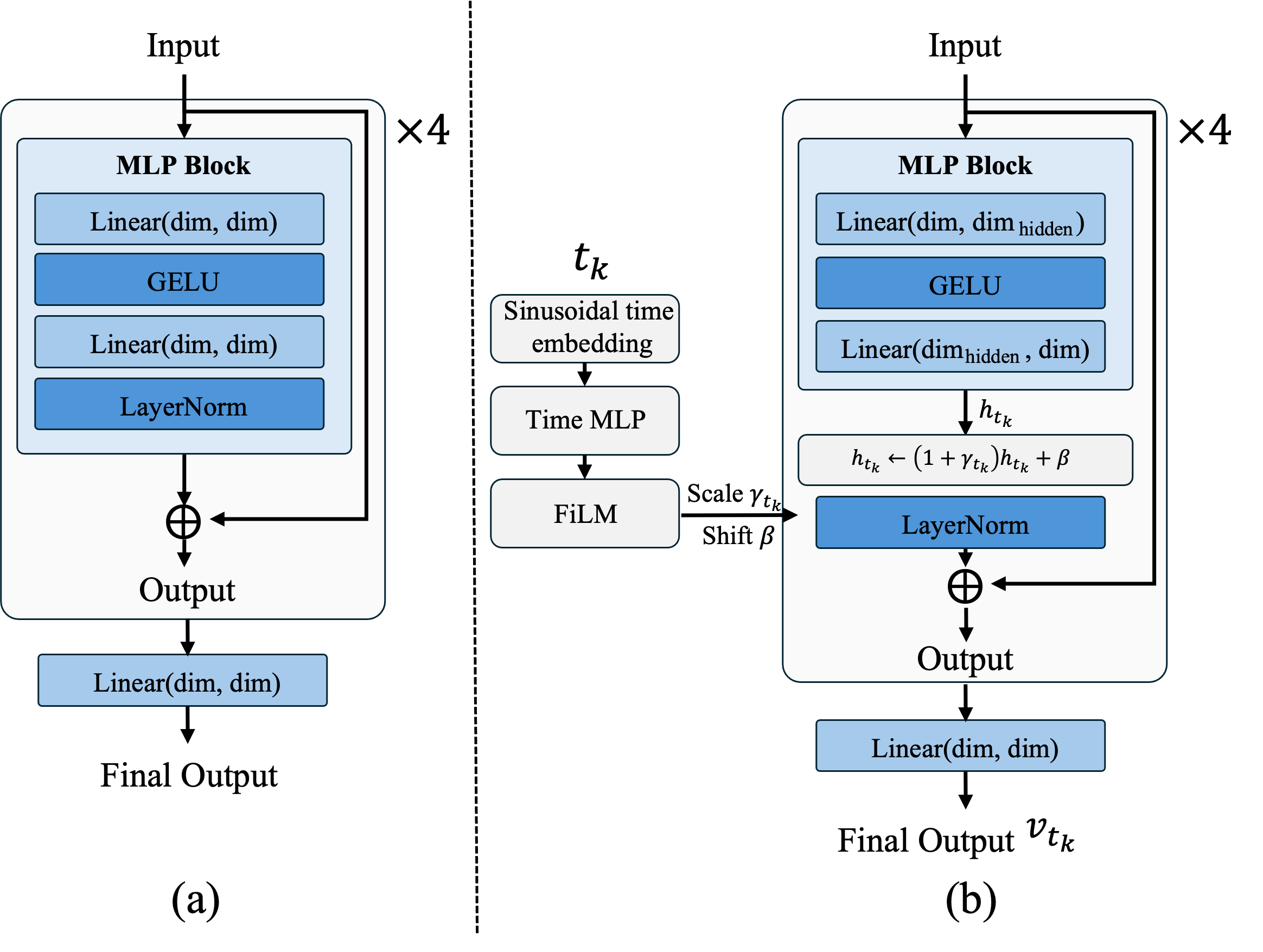}
   \caption{Illustration of network architectures. (a) MLP baseline, (b) Velocity network of DINOde.}
   \label{fig:supple_mlp-vs-ours}
   \vspace{-5pt}
\end{figure}

As shown in Fig.~\ref{fig:supple_mlp-vs-ours}, throughout the experiments, the MLP baseline and our velocity network are designed to be identical except for the time conditioning to ensure a fair comparison. To minimize the parameter increase introduced by the Time MLP and FiLM layers used only in the velocity network, we set $dim = 1024$ and $dim_{hidden}$ $= 512$ for the velocity network. As a result, the MLP baseline has 8.4M parameters, while the velocity network in DINOde has 7.5M parameters (about 10\% \underline{\textbf{fewer}} than the MLP baseline).

\begin{table*}[t]
\centering
\caption{Background categories used for each benchmark, following the CLIPer~\cite{sun2025cliper}.}
\label{tb:bg_class}
\resizebox{0.99\textwidth}{!}{
\begin{tabular}{c|c}
\toprule
\textbf{Benchmark} & \textbf{Background Classes} \\
\midrule
\multirow{2}{*}{\textbf{V21}} & ground, land, grass, tree, building, wall, sky, lake, water, river, sea, railway, railroad, \\
& keyboard, helmet, cloud, house, mountain, ocean, road, rock, street, valley, bridge, sign \\
\midrule
\multirow{2}{*}{\textbf{Object}} & ground, land, grass, tree, building, wall, sky, lake, water, river, sea, railway, \\
& railroad, helmet, cloud, house, mountain, ocean, road, rock, street, valley, bridge \\
\midrule
\textbf{C60} & background \\
\bottomrule
\end{tabular}
}
\end{table*}

\section{\textit{Additional Discussion}}

\paragraph{\textbf{Computational Cost.}}
Table~\ref{tab:comp_cost} reports the inference time of the main
components in our pipeline, including the visual backbone,
the CLIP text encoder, and the visual-text alignment.
All measurements are averaged across the VOC dataset.
Here, we compare a discrete single-step projection (MLP) with the proposed Semantic Text Flow (STF) evaluated using different numbers of Euler integration steps ($N_{\text{step}}$).
The standard MLP baseline requires 1.05 ms, whereas performing a single forward integration step of STF ($N_{\text{step}}=1$) takes 2.01 ms.
Extending the continuous transformation trajectory to
$N_{\text{step}}=10$ results in an inference time of 17.70 ms.
When compared with the computational cost of the visual backbone
(45.75 ms) and the CLIP text encoder (354.49 ms),
the additional overhead introduced by the 10-step numerical flow
remains relatively small.
Considering the performance improvement achieved by DINOde,
this additional cost represents a modest trade-off between
accuracy and computational overhead. Since the proposed DINOde is structurally almost similar to the MLP baseline, the GPU consumption remains relatively the same. 
For a fixed class set, this alignment cost is incurred only once and cached, so the per-image inference speed remains comparable to prior work: DINOde runs at 22.2 FPS versus 22.3 FPS for Talk2DINO~\cite{talk2dino}, with a one-time text-anchor setup of 381.7 ms versus 336.0 ms.

\begin{table}[t]
\centering
\small
\caption{Inference time analysis evaluated on the Pascal VOC dataset.}
\label{tab:comp_cost}
\resizebox{0.5\columnwidth}{!}{
\begin{tabular}{l|cc}
\toprule
\textbf{Method} & \textbf{Time (ms)} & \textbf{VRAM (MB)}\\
\midrule
MLP & 1.05 & 82.4\\
STF ($N_{step}=1$) & 2.01 & 93.8\\
STF ($N_{step}=10$) & 17.70 & 93.8 \\
\midrule
Visual Backbone & 45.75 & 1208.7\\
Text Encoding & 354.49 & 1693.4\\

\bottomrule
\end{tabular}
}
\end{table}

\begin{figure}[t]
  \centering
   \includegraphics[width=0.70\linewidth]{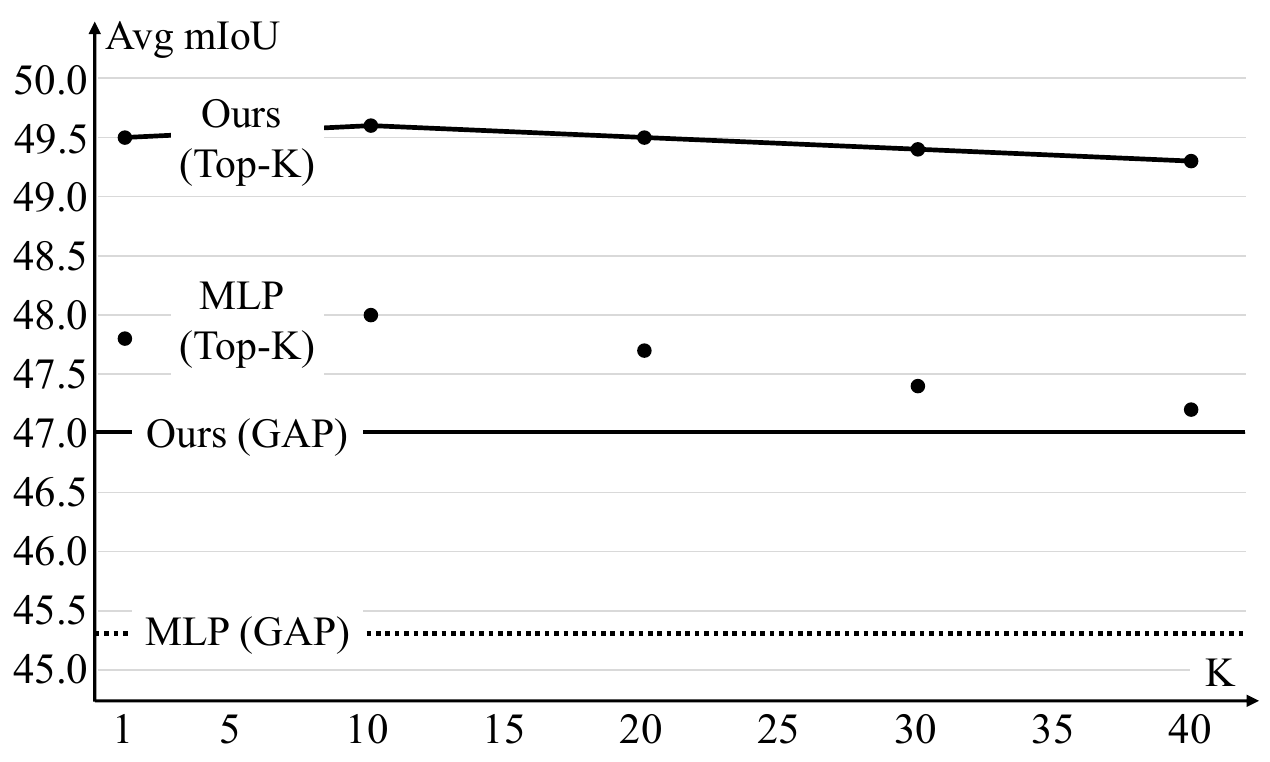}
   \caption{Ablation study on the pooling strategy and hyperparameter $K$.}
   \label{fig:supple_k_abl}
\end{figure}

\paragraph{\textbf{Ablation Study on Pooling.}}
Regarding the pooling function ``$\text{Pool}$'' introduced in Eq.~\textcolor{eccvblue}{6} of the main paper, we conducted an ablation study to evaluate the impact of different pooling strategies and the hyperparameter $K$ in Min-Max Top-K pooling~\cite{yoon2022adversarial}, as illustrated in Fig.~\ref{fig:supple_k_abl}. Performance is reported as the average mIoU across eight benchmarks.
First, our ODE-based method consistently surpasses the MLP-based baseline regardless of the pooling strategy, the Global Average Pooling (GAP) or Min-Max Top-K Pooling (Top-K). This performance gap clearly demonstrates the superiority of our continuous flow-based alignment, which more effectively bridges the modality gap than the baseline.
Next, we compared Top-K pooling with GAP. As illustrated in Fig.~\ref{fig:supple_k_abl}, Top-K pooling consistently outperforms GAP across all $K$ configurations, irrespective of whether an MLP-based or the proposed ODE-based alignment is employed. This trend can be attributed to the architectural characteristics of DINOv3, which generates highly fine-grained and well-aligned patch-level features. 
While GAP suffers from information dilution by averaging all patch tokens, including ambiguous elements, Min-Max Top-K pooling is likely to facilitate a more robust manifold alignment by selectively aggregating discriminative representations. 
Lastly, we conducted experiments of ablation on $K$, while performance remained relatively stable across the tested range.
Although the peak performance was recorded at $K=10$, we selected $K=20$ to prioritize overall training stability and the robustness of feature representation.

\paragraph{\textbf{Effect of Solver.}}
Table~\ref{tab:supp_solver} compares Euler integration, which is used in our default configuration, with the second-order Runge--Kutta (RK2)-Heun and RK2-Midpoint solvers and the fourth-order Runge--Kutta (RK4) solver.
Interestingly, all solvers achieve nearly identical average mIoU of around 49.5\% across the eight benchmark datasets, with only marginal variations on individual datasets.
This suggests that the learned velocity field produces a sufficiently smooth trajectory, for which higher-order numerical integration does not provide a noticeable performance benefit.
Considering both performance and computational efficiency, we adopt Euler integration as our default solver.

\begin{table*}[t]
\centering
\caption{Quantitative comparison of Euler's method, RK2 (Heun, Midpoint), and RK4 ODE solvers across eight benchmarks.}
\vspace{-6pt}
\label{tab:supp_solver}
\resizebox{0.90\textwidth}{!}{
\begin{tabular}{c|cccccccc|c}
\toprule
\textbf{ODE Integration} & \textbf{V20} & \textbf{C59} & \textbf{Stuff} & \textbf{City} & \textbf{ADE} & \textbf{V21} & \textbf{C60} & \textbf{Object} & \textbf{Avg} \\
\midrule
Euler's Method & 91.1 & 44.5 & 31.6 & 45.4 & 25.4 & 69.2 & 40.7 & 48.0 & 49.5 \\ \hline
RK2-Heun       & 91.0 & 44.1 & 31.9 & 46.1 & 25.4 & 69.0 & 40.3 & 48.5 & 49.5 \\ \hline
RK2-Midpoint   & 91.0 & 44.1 & 31.9 & 46.1 & 25.4 & 69.1 & 40.2 & 48.6 & 49.5 \\ \hline
RK4 & 90.9 & 44.1 & 31.9 & 45.9 & 25.3 & 68.7 & 40.1 & 48.4 & 49.5 \\

\bottomrule
\end{tabular}
}
\end{table*}

\paragraph{\textbf{Comparison with PEFT Methods.}}
Attaching a trainable projector on top of the frozen text encoder can be viewed as a form of parameter-efficient fine-tuning (PEFT). To examine whether the gains of DINOde stem from continuous alignment rather than parameter-efficient adaptation, we compare against LoRA, OFT, and COFT applied to the CLIP text encoder under a matched number of trainable parameters (Table~\ref{tab:supp_peft}). While their per-dataset performance varies, the PEFT baselines achieve similar averaged mIoU. DINOde consistently outperforms all PEFT baselines as well as the MLP, confirming that the improvement originates from the ODE-based cross-modal alignment itself.

\begin{table}[t]
\centering
\caption{Comparison with PEFT-based baselines applied to the CLIP text
encoder under a matched parameter budget.}
\label{tab:supp_peft}
\begin{tabular}{lccccc}
\toprule
Method & \textbf{LoRA} &\textbf{OFT} & \textbf{COFT} & \textbf{MLP} & \textbf{Ours} \\ 
\midrule
Avg.\ mIoU $\uparrow$ & 46.13 & 46.14 & 46.12 & 47.65 & \textbf{49.50} \\
\bottomrule
\end{tabular}
\end{table}

\paragraph{\textbf{Generalization to Non-DINO Backbones.}}
Beyond the DINOv3 backbone scaling studied in the main paper (Table~3), we further evaluate whether the continuous alignment generalizes to other visual backbones, including SAM~\cite{sam1}, SigLIP~\cite{siglip}, iBOT~\cite{zhou2021ibot}, and V-JEPA 2.1~\cite{vjepa21} (Table~\ref{tab:supp_backbone_generalization}). DINOde surpasses the MLP baseline across all backbones. For SAM, whose representations are not optimized for
class-level semantics, both methods yield low absolute performance, yet DINOde still improves over the MLP. These results indicate that the benefit of ODE-based alignment is not specific to DINO but holds across visual backbones with differing semantic structure.

\begin{table}[t]
\centering
\caption{Generalization to non-DINO visual backbones (Avg.\ mIoU).}
\vspace{-5pt}
\label{tab:supp_backbone_generalization}
\begin{tabular}{lcccc}
\toprule
\textbf{Align Method}\  & \textbf{SAM}~\cite{sam1} & \textbf{SigLIP}~\cite{siglip} & \textbf{iBOT}~\cite{zhou2021ibot} & \textbf{V-JEPA 2.1}~\cite{vjepa21} \\ 
\midrule
MLP  & 7.0 & 26.0 & 33.1 & 31.7 \\ \hline
Ours & \textbf{7.7} & \textbf{26.9} & \textbf{34.5} & \textbf{34.7} \\
\bottomrule
\end{tabular}
\end{table}

\begin{figure}[t]
  \centering
   \includegraphics[width=0.71\linewidth]{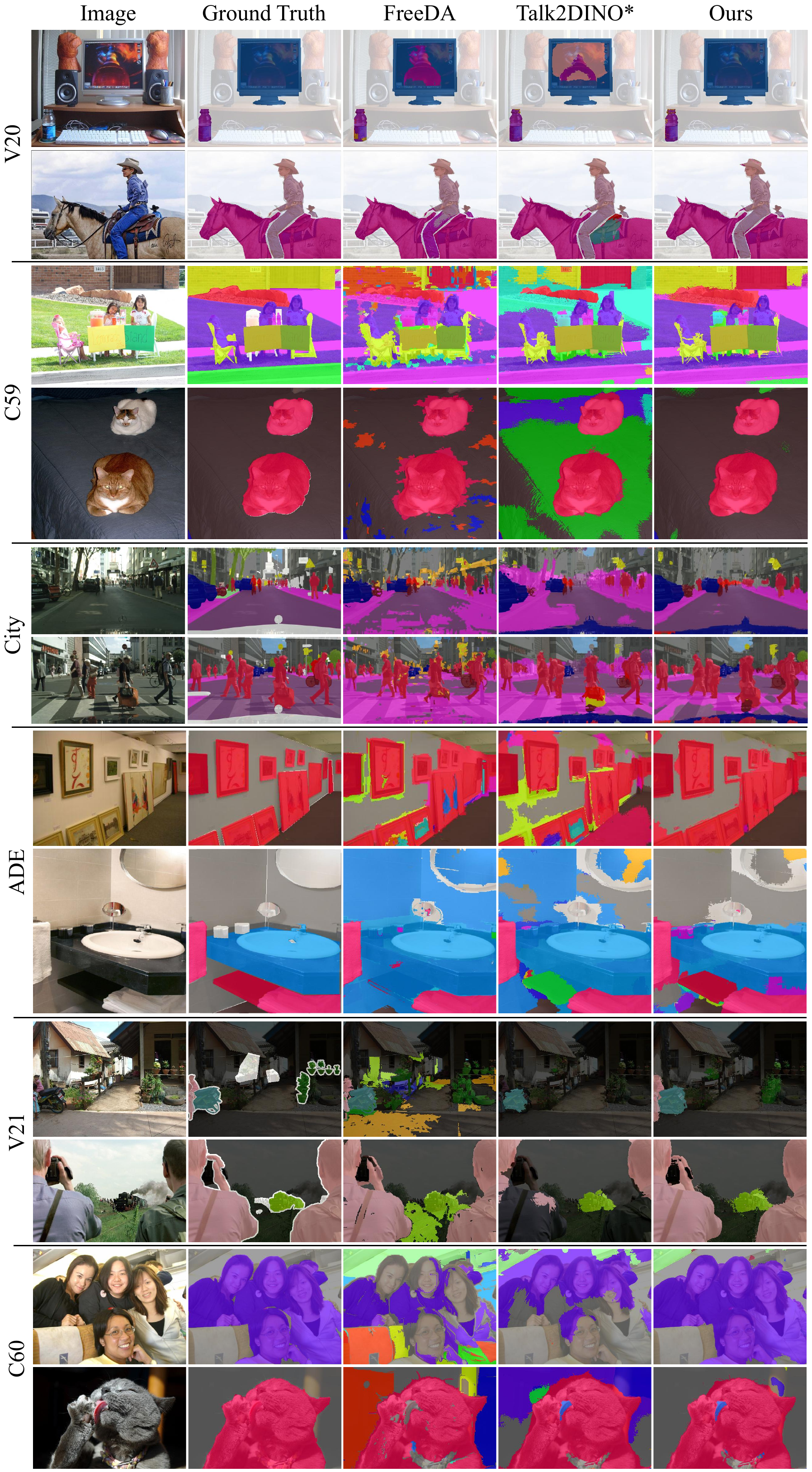}
   \vspace{-6pt}
   \caption{Additional Qualitative comparison with the state-of-the-art methods. The result of Talk2DINO* is obtained using the DINOv3 backbone for a fair comparison.}
   \label{fig:supple_qual}
\end{figure}

\begin{figure}[t]
  \centering
   \includegraphics[width=0.71\linewidth]{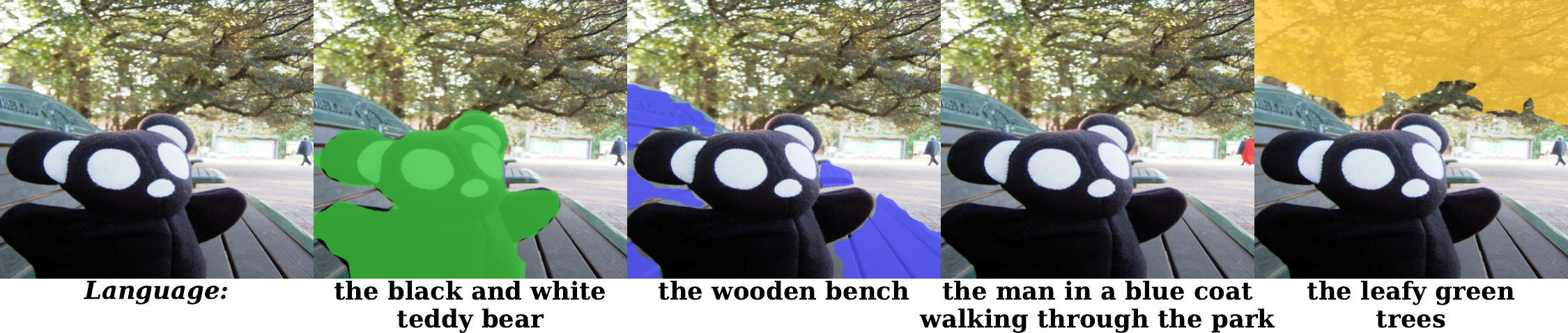}
   \vspace{-6pt}
   \caption{Qualitative results on RefCOCO+~\cite{refcocoplus}-style natural-language queries.}
   \label{fig:supp_nl_qual}
\end{figure}

\paragraph{\textbf{Additional Qualitative Results.}}
To further demonstrate the effectiveness of the proposed DINOde, we present additional qualitative comparisons in Fig.~\ref{fig:supple_qual}, complementing the qualitative results shown in the main paper. We compare FreeDA, a training-free approach, Talk2DINO* (w/ DINOv3 backbone), which similarly utilizes a self-supervised visual backbone, and the proposed method. Qualitative examples are collected from six datasets: V20, C59, City, ADE, V21, and C60.
As illustrated in the figure, while previous state-of-the-art methods often fail to produce accurate segmentation masks, DINOde generates predictions that closely resemble the ground truth. This observation suggests that the proposed continuous alignment effectively aligns the self-supervised visual features with the text embeddings. Furthermore, as observed on City and ADE, where previous methods typically show relatively lower performance, DINOde still produces promising segmentation results, further demonstrating its effectiveness.
To further probe behavior beyond fixed prompt templates, we evaluate DINOde on RefCOCO+~\cite{refcocoplus}-style natural-language
queries (Fig.~\ref{fig:supp_nl_qual}). DINOde successfully localizes the target regions described by free-form expressions, suggesting that the aligned manifold generalizes to richer language beyond fixed class names.

%
%

\end{document}